\documentclass[conference]{IEEEtran}
\IEEEoverridecommandlockouts
\usepackage{cite}
\usepackage{amsmath,amssymb,amsfonts}
\usepackage{algorithmic}
\usepackage{graphicx}
\usepackage{textcomp}
\usepackage{xcolor}
\def\BibTeX{{\rm B\kern-.05em{\sc i\kern-.025em b}\kern-.08em
    T\kern-.1667em\lower.7ex\hbox{E}\kern-.125emX}}

\newcommand{\eeric}[1]{}
\newcommand{\ebb}[1]{{#1}}

\newcommand{\eric}[1]{}
\newcommand{\com}[1]{}
\newcommand{\bb}[1]{{#1}}

\begin{document}

%\title{Investigating Imitation Methods for End-to-End Autonomous Driving on Urban Environments}
\title{Generative Adversarial Imitation Learning for End-to-End Autonomous Driving on Urban Environments}

% PROPOSTA DE TITULO:
% On Generative Adversarial Imitation Learning for End-to-End Autonomous Driving on Urban Environments

\author{\IEEEauthorblockN{Gustavo Claudio Karl Couto}
\IEEEauthorblockA{\textit{Automation and Systems Engineering Department} \\
\textit{Federal University of Santa Catarina}\\
Florianopolis, Brazil \\
gustavo.karl.couto@posgrad.ufsc.br}
\and
\IEEEauthorblockN{Eric Aislan Antonelo}
\IEEEauthorblockA{\textit{Automation and Systems Engineering Department} \\
\textit{Federal University of Santa Catarina}\\
Florianopolis, Brazil \\
eric.antonelo@ufsc.br}
}

\maketitle

\begin{abstract}
\bb{Autonomous driving is a complex task, which has been tackled since the first self-driving car ALVINN in 1989, with a supervised learning approach, or behavioral cloning (BC). In BC, a neural network is trained with state-action pairs that constitute the training set made by an expert, i.e., a human driver. However, this type of imitation learning does not take into account the temporal dependencies that might exist between actions taken in different moments of a navigation trajectory. These type of tasks are better handled by reinforcement learning (RL) algorithms, which 
need to define a reward function.
On the other hand, more recent approaches to imitation learning, such as Generative Adversarial Imitation Learning (GAIL), can train policies without explicitly requiring to define a reward function, allowing an agent to learn by trial and error directly on a training set of expert trajectories. 
In this work, we propose two variations of GAIL for autonomous navigation of a vehicle in the realistic CARLA simulation environment for urban scenarios. Both of them use the same network architecture, which process high-dimensional image input from three frontal cameras, and other nine continuous inputs representing the velocity, the next point from the sparse trajectory and a high-level driving command. We show that both of them are capable of imitating the expert trajectory from start to end after training ends, but the GAIL loss function that is augmented with BC outperforms the former in terms of convergence time and training stability.}
%
%Imitation learning for time-dependent tasks is a novel field of robotics, in which an agent learns to perform a task using demonstrations from experts. It's a generic learning method that's especially interesting for autonomous driving in urban environments where an agent is expected to comply with explicit rules and implicit norms to not generate noise in the environment. This paper evaluates a Generative Adversarial Imitation Learning architecture with improved robustness for autonomous end-to-end driving in a realistic urban environment.
\end{abstract}


% GAIL + simulador CARLA
% bc GAIL > GAIL
% no pretraining
% GAIL > BC ambiente longo
% 3 cameras x3 RGB, +9 continuous input (velocidade, proximo ponto na ref., comando)

% \eric{---Everything in RED, for Gustavo to check/answer...}\\
% \\
% \eric{TODO:
% - figura arquitetura com entradas de image + continuous input \\
% - tentar rodar mais simulações/repetições\\
% - linha horizontal BC \\
% - adaptar seção de GAIL de dropout para BC GAIL (me avisar quando fizer) \\
% ----- Verificar e fazer/responder todos TODO ao longo do texto que eu indiquei.
% }
% \\
% \\
% \com{For Eric: Abstract, introduction, GAIL section, and review overall text }

% minimal environment interactions

\section{Introduction}
\label{introduction}
%%%%%%%%%%%%%%%%%%%%%%%%% Start Eric's text
Imitation learning is an approach whereby a model is created to imitate an expert by training on a fixed set of observation-action samples (or trajectories) obtained from that expert. This happens without the possiblity of querying the expert while training.
Behavioral cloning (BC) \cite{Pomerleau1991,Antonelo2014,Antonelo2010} is one approach for imitation learning that relies on supervised learning to learn a mapping between observations and actions. It has been used for autonomous navigation since 1989, starting with the self-driving car ALVINN \cite{Pomerleau1991} that relied on camera images as input to a neural network (NN) to drive the car.  
BC has the issue of sample complexity since it requires a lot of training data (observation-action samples) generated by experts (e.g. human drivers) to work well in practice. However, BC will always suffer from {cascading errors} and covariate shift \cite{Ross2010} since their models are trained only on a subset of the necessary samples (observation-action pairs) for safe, robust driving: as soon as the self-driving car encounters a new road and starts shifting slightly towards the left or right side of the lane, it will feedback its mistake through new observations fed to the NN, which in turn will shift even more the car until no valid action can be taken anymore.

On the other hand, policies learned by {Reinforcement Learning} (RL) solve the issue of cascading error since they learn from information of whole sample trajectories and not just isolated observation-action samples as in BC, but require a reward (cost) function to be defined for finding the optimal policy. In RL, training is an evolutionary method where an agent learns by trial and error, i.e., interacting with the environment, and receiving a reward signal indicating the quality of the solution found.

In the context of imitation learning, RL can be used to learn to imitate expert driving trajectories in a process called
%Given that we have a collection of expert driving trajectories as training dataset (much like in BC),
inverse reinforcement learning (IRL) \cite{Ng2000,Russell1998}. IRL can be used to find driving policies by:
first finding a cost function under which the expert, i.e., the set of training trajectories, is uniquely optimal; and then using RL algorithms that optimize the learned cost function. 
IRL is usually expensive to run since it requires RL in an inner loop, and thus, has difficulties in scaling to large environments. Recent work in IRL seeks to deal with these issues \cite{Finn2016,Levine2012}. Still, learning a cost function in IRL makes the problem more computationally expensive than just learning a policy directly from the training trajectories. 

One of the recently developed sample refiefficient approaches for imitation learning comes from {Generative Adversarial Imitation Learning} (GAIL) framework \cite{Ho2016}, which 
can actually generate policies directly from the expert trajectories without having to learn any cost function as in IRL, and is scalable to relatively large environments, such as the one in autonomous driving. 
However, training in GAIL can be unstable and difficult to achieve a satisfactory result depending on the task. While it is sample efficient in terms of the required number of expert trajectories, it is not so efficient in the number of environment interactions needed for convergence.

On the other hand, BC converges in a few epochs, but assumes that its dataset is composed of independent and identically distributed samples. A recently developed approach \cite{jena2020augmenting} combines both BC and GAIL losses into an integrated loss function for stable and sample efficient imitation learning. They have evaluated it on low-dimensional control tasks, and also on the high-dimensional image-based task of CarRacing from OpenAI Gym.

In our work, we propose a GAIL-based architecture for end-to-end autonomous urban navigation, which is evaluated on fixed trajectories  in the realistic autonomous driving CARLA simulator \cite{dosovitskiy2017carla}. 
The agent receives a high-dimensional image input from three frontal cameras, as well as other continuous inputs such as velocity and next point of a sparse GPS trajectory in the local vehicle's frame of reference. 
As far as the authors know, this is the first proposal of architectures based on conventional GAIL and GAIL augmented with BC \cite{jena2020augmenting} for end-to-end imitation learning in the CARLA simulator, also considering a much higher observation space (in relation to the simpler CarRacing environment, for instance).
Our experiments have shown that although the GAIL architecture can learn to imitate the expert well, GAIL augmented with BC has much faster convergence to the desired navigation trajectory.

In Section \ref{related_work}, we give a brief overview on some related works. Next, we present the main methods such as GAIL and GAIL augmented with BC as well as the agent architecture in Section \ref{methods}.
Section \ref{experiments} describes the experiments, datasets and settings, while the results are presented in Section \ref{results}. Conclusions and future work are drawn in Section \ref{conclusion}.

\section{Related Work}
\label{related_work}
% \eric{TODO: For each cited related work: describe succintly: observation space; action space; method; results; and, if relevant, the main difference to our approach}

% \eric{TODO: check the way I organized the subsections; and reorganize the references adequately if necessary.}

%-EA: this is not part of related work (commented)
%CARLA simulator \cite{dosovitskiy2017carla} is an open-source platform created by a partnership of a university, research center, and a private company to develop a safe, and realistic environment to conduct experiments of autonomous driving in urban environments.

%-EA: this is not part of related work (commented)
%Since CARLA was released, it has allowed the development of end-to-end deep learning algorithms, that learn a policy control interacting directly with an environment. The policy receives inputs from sensory inputs as cameras inserted on the vehicle and generates the steering, throttle, and brake outputs that compose the driving signal.

\subsection{Behavior cloning}
\eeric{In general, you may add other references in each section as "Other related work can be found in ..."}

% One of the most important recent works on imitation learning for end-to-end autonomous driving on urban environments was a conditional imitation learning algorithm \cite{codevilla2018endtoend}, that performs a conditional behavior clone, a type of supervised imitation learning algorithm augmented with a command signal that describes the trajectory to be followed by the agent.

\ebb{One of the most important recent works on imitation learning for end-to-end autonomous driving on urban environments corresponds to a conditional imitation learning algorithm implemented in \cite{codevilla2018endtoend}.
Based on behavior cloning, the learning is conditioned on a high-level command signal that indicates the way through the trajectory to be followed by the agent.
}
\eric{question: what sort of command? right, left? or next point?}
%Answer: We refer to this architecture as branched. The branches Ai are forced to learn sub-policies that correspond to different commands. In a driving scenario, one module might specialize in lane following, another in right turns, and a third in left turns. All modules share the perception stream.
% The observation space of the algorithm consists of three images from frontal cameras 
% \eeric{images from three frontal cameras?}
% installed on the vehicle, the vehicle speed if available, and a high level command for the trajectory control or the vector to the goal on the vehicle coordinates. The action space consists of a two dimensional vector containing steering angle and acceleration. The experiments are conducted both on CARLA and on a physical system, the last is an off-the-shelf 1/5 scale truck, modified to add three cameras and a single board computer for autonomous driving.
The observation space consists of: images from three frontal cameras
installed on the vehicle; 
the vehicle speed if available;
and a high-level driving command. 
%or the vector to the goal \eeric{***} on the vehicle coordinates. 
\ebb{The action space consists of the vehicle's steering angle and acceleration. 
The experiments are conducted both on CARLA and on a real setting with an off-the-shelf 1/5 scale truck, which was adapted with three frontal cameras and a single board computer for autonomous driving.}

% Both experiments were successful and also generalized well, so that the work can be considered a milestone, adding the capability to the agent to follow generic trajectories, using the conditional learning, where most of the works before concentrated on fixed paths. The work was also important as proof that an agent developed on CARLA, is also capable of being deployed in the real world.
\ebb{Both experiments were successful and also generalized well, making it
a milestone for enabling an agent to learn to follow generic trajectories, whereas most of previous works have concentrated on fixed paths. The work has shown that
CARLA serves as an important platform to analyse agents and learning approaches before deploying them to the real world.
}

\eeric{What about other references? Can you just cite others?
GAIL is also imitation learning. Should the title be "behavior cloning"?
cite{codevilla2019exploring}
esse eh uma sequencia do primeiro trabalho, explorando os limites da arquitetura bc pro carla, e propondo uma um benchmark unificado pra comparar agentes treinados com BC
Sim o titulo behivior cloning seria mais preciso
}

%The limitations of the technique were explored in another work \cite{codevilla2019exploring}, that built a framework to evaluate and compare different behavior cloning approaches, and consolidated the knowledge from that technique.

%-EA: this is not part of related work (commented)
%Behavior cloning as a supervised learning technique assumes that samples are independent and identically distributed and suffers from co-variance shifts between inputs from demonstrations and a real-time evaluation \cite{ross2011reduction} \cite{efficientReductions}. So depends on large data set and has a performance limitation. To not make this assumption it is imperative to make use of reinforcement learning technique, that interacts with the environment during the learning phase and learns the time dependency of actions.

\subsection{Reinforcement learning}
\eeric{revisado. revisado 2.}

% As an approach for reinforcement on end-to-end autonomous driving, the Controllable Imitative Reinforcement Learning \cite{cirl2018} was proposed as a two-phase algorithm that pre trains a policy using behavior cloning and then refines that policy during interaction with the environment using engineered reward functions to enforce the best behavior of the agent.
The Controllable Imitative Reinforcement Learning \cite{cirl2018} was proposed as a two-phase algorithm that pre-trains a policy using behavior cloning and then refines it during interaction with the environment using an engineered reward function to enforce the best behavior of the agent. This reward signal is composed by negative rewards for 
abnormal steer angles, damage from collisions, going over the sidewalk or the opposite lane, and by a positive reward for reaching a desired speed.
% \eeric{Can you give more details on this?
% What about:
% The work implements a reward module that return a sum of five terms based on the policy action and the environment state: $r_s$ a negative reward to abnormal steer angles, like an angle to the opposite direction during a turn;$r_v$ a positive reward for speed up to a desired speed;$r_d$ a negative reward for damage from collisions; $r_r$ and $r_o$ negative rewards for overlap with sidewalk and opposite lane respectively.}

% The observation space of the algorithm consists on images from a front facing camera, the vehicle speed and four options of high level command ("follow-lane", "turn left", "turn right", "go straight"). The image from the camera feeds the convolution layers from the actor critic network while the speed enters ahead on the fully connected part of the network. The high level command is used to select the final branch of the network, that outputs the signals for the steering wheel, throttle, and brake.
The observation space of the algorithm consists of an image from a front-facing camera, the vehicle speed and four options of high-level commands ("follow-lane", "turn left", "turn right", "go straight"). 
The image from the camera feeds the convolution layers of the actor-critic network while the speed is fed directly to the fully connected layers of the network. The high-level command is used to select the final branch of the network, that outputs the signals for the steering wheel, throttle, and brake.
%Those reward functions are hard to engineer for a complex task like autonomous driving and may lead the agent to unnatural or unfriendly behavior in the traffic.

%%%%%%%%%%% CARLA related works
\subsection{Apprenticeship Learning}
\eeric{revisado; 
TODO: Change huang2021learning reference for 2020 CoRL conference
done}

% One of the last works to break one mark on CARLA benchmark was \cite{huang2021learning}, a multi stage approach that on the first stage trains a teacher using behavior cloning with CARLA privileged information about the landscape and other agents on its observation space. On the second stage it trains a vision based agent without access to privileged information using apprenticeship learning.

% One of the last works to break one mark on CARLA benchmark was \cite{huang2021learning},
In \cite{chen2019learning}, 
a multi-stage learning approach is employed that succeeded in the CARLA benchmark \cite{leaderboard_2020}. 
\eeric{cite benchmark}
Their method is based on the training of a teacher in a first stage using behavior cloning, which has privileged information about the landscape and other agents on its observation space in the CARLA simulator. 
In the second stage, a vision-based agent is trained without access to privileged information using apprenticeship learning \cite{apprenticeshiplearning}.
\eeric{cite apprenticeship: cited}.

% The observation space of the final networks from that work, consists of a 384x184 RGB image from a front facing camera and the vehicle velocity. The action space generated by the network consists of four points generated by network's four heads representing four high level commands ("follow-lane", "turn left", "turn right", "go straight").
The observation space of the agent consists of a 384x184 RGB image from a front-facing camera and the vehicle velocity. 
The action space corresponds to $K$ waypoints representing the agent's future locations on the next $K$ states in the agent reference. The waypoints are generated by network's four heads representing the following high-level commands "follow-lane", "turn left", "turn right", "go straight".
%The action space correspond to $K$ waypoints generated by network's four heads representing the following high-level commands "follow-lane", "turn left", "turn right", "go straight".
\eeric{What is this point? Is it a point from a GPS trajectory? How it is defined? Is it supervised learning?
}

A low-level rule-based controller uses those waypoints predicted by the network and the high-level command given to the vehicle to generate the car attitude control (steer, throttle, brake).

\subsection{GAIL for autonomous driving}
\eeric{revisado; 
}

%From the related works in the field, the one that is most close to this work is InfoGail \cite{li2017infogail}. 
GAIL was first applied to autonomous driving in  \cite{li2017infogail}. 
In that work, a Wasserstein Gail is designed to control a vehicle on TORCS \cite{wymann2000torcs}, an open-source racing car simulator. The observation space consists of images taken from the front of the car, and some auxiliary information (the car velocity, the last two actions, and the damage to the car). 
The action space corresponds to a three-dimensional vector with the steering command, acceleration, and brake. 

Their method, called InfoGAIL, still augments the standard GAIL with a replay buffer and a reward signal with constant reward to encourage the agent to stay alive. 
The focus of InfoGAIL is to demonstrate a capability to learn a policy that can switch between driving behaviors by disentangling in an unsupervised way the different modes of behavior present in the expert's demonstrations. 
%The and train a controllable policy on those different modes of behaviour.
\bb{Other related work can be found in \cite{huang2021learning,huang2021efficient,zhou2020smarts}}.
%
%Another GAIL evaluation on autonomous driving was performed \cite{huang2021learning} that uses a GAIL on the top of a pipeline to train a high-level decision module of a pipeline controller for an autonomous vehicle on CARLA. \eric{Explicar melhor isso, e a diferença com o trabalho atual; é end-to-end?}

%\subsection{Imitation learning in other simulators}

%A related work evaluates the performance of a neural network controller on small trajectories \cite{huang2021efficient}, the controller is trained using deep reinforcement learning and regularized using the KL divergence to a policy trained using behavior cloning. \eric{not very clear; KL between what?}
%The evaluation is done using Smarts (Scalable multi-agent reinforcement learning training school for autonomous driving) \cite{zhou2020smarts}, a simulator for autonomous driving and the input to the agent is a semantic image of the environment. \eric{What is a semantic image? Is it only image, or are there commands, etc.?}

To our knowledge, this is the first time that GAIL or GAIL augmented with BC are evaluated for an end-to-end self-driving task in a highly realistic urban vehicle simulator as CARLA.

\section{Methods}
\label{methods}

\subsection{Generative Adversarial Imitation Learning}

\bb{In Generative Adversarial Imitation Learning (GAIL) \cite{Ho2016}, basically, there are two components that are trained iteratively in a min-max game: a discriminative classifier $D$ is trained to distinguish between samples generated by the learning policy $\pi$ and samples generated by the expert policy $\pi_E$ (i.e., the labelled training set); and the learning policy $\pi$ is optimized to imitate the expert policy $\pi_E$. Thus, in this game, both $D$ and $\pi$ have opposite interests: $D$ feeds on state-action pair $(s,a)$ and its output seeks to detect whether $(s,a)$ comes from learning policy $\pi$ or expert policy $\pi_E$; and $\pi$ maps state $s$ to a probability distribution over actions $a$, learning this mapping by relying on $D$'s judgements on state-action samples (i.e., $D$ informs how close $\pi$ is from $\pi_E$).
Mathematically, GAIL finds a saddle point $(\pi,D)$ of the expression:}
\begin{equation}
\mathop{\mathbb{E}}_\pi[\log(D(s,a))] +  \mathop{\mathbb{E}}_{\pi_E}[\log(1-D(s,a))]  - \lambda H(\pi) 
\label{eq:gail}
\end{equation}
\bb{
where $D: S \times A \rightarrow  (0,1) $, $S$ is the state space, $A$ is the action space; $\pi_E$ is the expert policy; $H(\pi) $ is a policy regularizer controlled by  $\lambda >= 0$ \cite{Bloem2014}.
GAIL works similarly to generative adversarial nets (GANs) \cite{Goodfellow2014}, which was first used to learn generators of natural images. Both $D$ and $\pi$ can be represented by deep neural networks. In practice, a training iteration for $D$ uses Adam gradient-based optimization \cite{Kingma2014} to increase (\ref{eq:gail}), and in the next iteration, $\pi$ is trained with any on-policy gradient method such as Proximal Policy Optimization (PPO) \cite{ppoOriginal}
%Trust-Region Policy Optimization (TRPO) \cite{Schulman2015} 
to decrease (\ref{eq:gail}).}

%Generative Adversarial Imitation Learning is a method inspired by Adversarial Networks, where two training is a game of min-max between both networks. Where the policy tries to minimize the Wasserstein distance representing the distance of the state action distribution generated by the policy to the state action distributions from the experts, while the critic is trying to maximize that difference.

\subsection{GAIL and BC augmentation}
\subsubsection{Wasserstein loss}

\bb{Instead of the original loss function of GAIL, as in (\ref{eq:gail}), in this work, we employ}
its improved version using the Wasserstein distance between the policy distribution $P\tau_{\pi}$ and expert distribution $P\tau_E$ \bb{as loss function for training the discriminator, as in \cite{Zhang_2020, li2017infogail}.}

The Wasserstein distance measures the minimum effort to move one distribution to the place of the other and gives a better feedback signal than the Jensen-Shannon divergence.
%that comes from treating the discriminator as a classifier, and gives no feedback signal, zero gradient, when both distributions don't share an overlapping region.
%EA: difficult to review this now; commented

The new loss function for the improved GAIL is:
%  \eric{Gustavo, cite here other GAIL with Wasserstein; Adapt notation to harmonize between sections A and B}
% \begin{equation}
% \mathop{\mathbb{E}}_{\pi_E}[D(s,a)] - \mathop{\mathbb{E}}_\pi[D(s,a)] - \lambda H(\pi)
% \label{eq:wgail}
% \end{equation}
% where the discriminator will try to increase (\ref{eq:wgail}), while $\pi$ seeks to minimize it.

% An additional condition for the discriminator $D$ converge to the Wasserstein distance is it to limit the discriminator network to the 1-Lipschitz functions space.
%

% \begin{equation}
% L_{gp} = (\left \| \bigtriangledown_{(\hat{s},\hat{a})\in P_{(\hat{s},\hat{a})}} D((\hat{s},\hat{a})) \right \|_2-1)^2
% \label{eq:gpenalty}
% \end{equation}

%The final loss function for the Wasserstein GAIL is:
\begin{equation}
\mathop{\mathbb{E}}_{\pi_E}[D(s,a)] - \mathop{\mathbb{E}}_\pi[D(s,a)] - \lambda H(\pi) - \lambda_2 L_{gp}
\label{eq:wgail2}
\end{equation}
where the discriminator will try to increase (\ref{eq:wgail2}), while $\pi$ seeks to minimize it; and $L_{gp}$ is a loss that penalizes the gradient
constraining the discriminator network to the 1-Lipschitz function space, according to \cite{gulrajani2017improved}.

\subsubsection{BC augmentation}

The behavior Cloning loss function can be defined as:

\begin{equation}
-\mathop{\mathbb{E}}_{\pi_E}[\log(\pi(a|s|))]
\label{eq:bc}
\end{equation}
which is the negative expectation of the log probability for the non deterministic policy generator to output the same actions as the expert on the same states from the expert dataset.

The BC augmentation is constructed taking a point from a line between the behavior cloning loss and the GAIL loss, as defined on the following equation:

\begin{equation}
\alpha L_{bc} + (1 - \alpha) L_{GAIL}
\label{eq:bcgail}
\end{equation}

On equation (\ref{eq:bcgail}), $L_{bc}$ is the behavior cloning loss function defined on (\ref{eq:bc}) and $L_{gail}$ is the GAIL loss function defined defined on (\ref{eq:wgail2}).

The $\alpha$ on (\ref{eq:bcgail}) controls the participation of each term during the training.  By the start of the GAIL training, the discriminator is yet not fully trained and the behavior cloning participation should be stronger. For that,  $\alpha$ should not be the same during the entire training and its value decreases during the training using a fixed decay factor.
This definition and the practical implementation follows  \cite{Jena2020}.

% The $L_{gp}$ is added to the 
%One more constraint id added to improve the response of the critic. This new constraint is called Consistency Term and uses the difference from the network response to points close to each other in the expert space, to guarantee a linear response from that region in the space.

%In practice, the consistency term is implement using points that are actually perturbations from the same expert sample generated by dropout layers in the critic.

%So in practice, the consistency term is implemented 

%$$CT|_{x_1,x_2}=E_{x\sim P_r}[max(0,d(D(x'),D(x''))+0.1\cdot d(D\_(x'),D\_(x''))-M']$$

%The adversarial imitation learning strategy can be defined as a
%mini max game:

%$$L = \mathbb{E}_{z\sim \mathbb{P}_z}[D(G(z))]-\mathbb{E}_{x\sim \mathbb{P}_r}[D(G(x))] + \lambda _1GP|_{\hat{x}}+\lambda _2CT|_{x',x''}$$

%And the reward input for the generator is defined by a reward shaping function as $e^x$ being $x$ the discriminator output.

\subsection{Agent and Network architecture}
\begin{figure*}[htp]
    \centering
    \includegraphics[width=5.5cm]{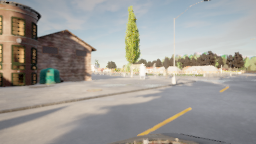}
    \includegraphics[width=5.5cm]{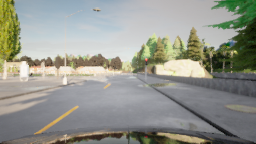}
    \includegraphics[width=5.5cm]{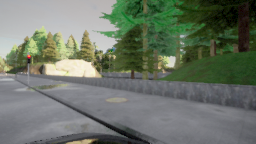}
    \caption{
    \bb{Images from the three frontal cameras located at the left, central, and right part of the vehicle, respectively. They were taken after the first few interactions of the agent in the CARLA simulation environment considering our defined trajectory. Each camera produces a RGB image with 144 pixels of height and 256 pixels of width.} 
    These images are fed to the networks as they are. \eric{TODO: verify this}. \eric{comment here the image resolution}
    }
    \label{fig:images}
\end{figure*}

\subsubsection{Agent}
% The experiment is inspired on the CARLA Leaderboard evaluation platform. The inputs for the agent come from sensors installed on the vehicle, three cameras on the front, an inertial unit generating data to calculate the vehicle linear speed and angular position, and a GPS unit for global position.
\bb{The autonomous car has several sensors, from which we consider: three frontal cameras (Fig.~\ref{fig:images}), an inertial unit used to compute the vehicle linear speed and angular position, and a GPS unit for global positioning.
}

% In the start of the episode the agent also receives the trajectory to perform, that is defined as vector of sparse points and high level commands defining the trajectory with no ambiguity.
\bb{
Before training begins, the agent has access to the whole trajectory it must perform, defined as a vector of sparse points and high-level driving commands that characterize the trajectory with no ambiguity.
These driving commands can be one from the following in this work:
\begin{itemize}
% \item CHANGE\_LANE\_LEFT: Move one lane to the left.
% \item CHANGE\_LANE\_RIGHT: Move one lane to the right.
\item LANE\_FOLLOW: Continue in the current lane.
\item LEFT: Turn left at the intersection.
\item RIGHT: Turn right at the intersection.
% \item STRAIGHT: Keep straight at the intersection.
\end{itemize}}

% On the trajectories from the experiment there are the commands of LEFT, RIGHT and LANE\_FOLLOW
% \eric{TODO: Comment in the text what commands are actually used in the experiment }

% The trajectory is fed to a route planner, that monitors the agent progress to generate the next trajectory point and high level command to the agent. 
\bb{Thus, the agent can use this trajectory to know which route the car should follow. In practice, this is accomplished by a route planner, that monitors the agent's progress and sends him the next target position in the car's frame of reference as well as
the high-level driving command. These two data, totalling 8 dimensions, are given as input to the agent. 
Notice that the command is input as an one-hot encoded vector.
}
\eric{verify above}
\eric{if possible, illustrate this with a figure}

\subsubsection{Networks architecture}

\begin{figure*}[htp]
    \centering
    \includegraphics[width=14cm]{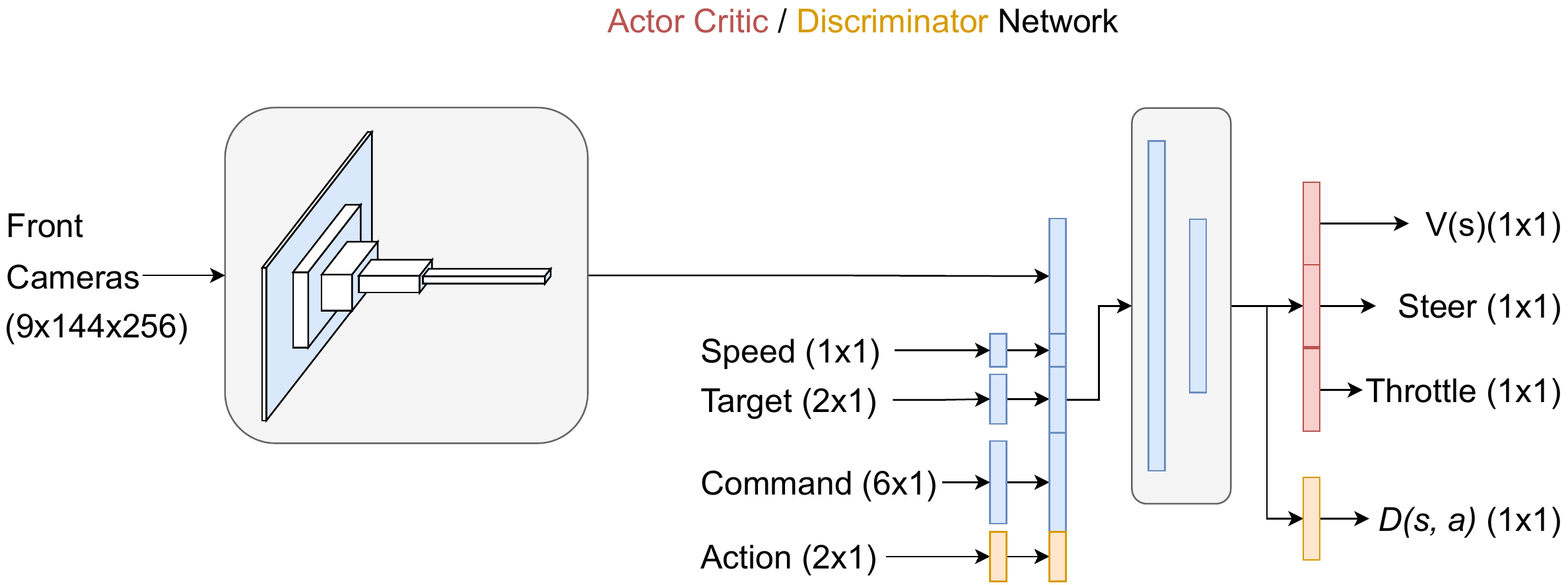}
    %EA: a fonte deve estar de tamanho parecido na fig. / redimensionei p/ isso
    \caption{
    \bb{Architecture of the actor-critic network and discriminator - each of them has its own separate network, with the latter having an additional input for the action, in orange color, and a sigmoidal output $D(s,a)$ instead of the output layer of the actor-critic network which consists of the 
    steering direction, throttle as actions for the actor (policy) and value of the current state $V(s)$ for the critic.
    The common, though not shared architecture (in blue) is
    composed of a convolutional block that process the images of the three frontal cameras, whose output features are concatenated with other nine continuous inputs for speed, next target point in the sparse GPS trajectory, and a high-level driving command.
    %and the agent action for the discriminator. 
    The resulting feature vector is input to a block of two fully-connected (FC) layers. 
    %The output layer consists of the steering direction, throttle as actions for the actor (policy), value of the current state $V(s)$ for the critic and a linear output for the discriminator.
    }
    \eric{check the above; what about means and stds of the gaussian distributions?}
    \eric{TODO: maybe mark the distinction between actor part and critic part, using different colors or ... ?}
    }
    \label{fig:nn}
\end{figure*}

The networks represented in Fig.~\ref{fig:nn} are composed of a convolutional block of four layers, with kernel size of 4 and stride of 2. Each layer in this block is followed by a leaky ReLU activation function, and the numbers of channels starts in 32 on the first layer and is multiplied by 2 on every new layer, ending with 256 channels.

That convolutional block is followed by a fully-connected network block with two layers, 
%between the layers there is an activation function of the type
with leaky ReLU activation function for the first hidden layer.
The second layer represents the output of the architecture.

Both actor-critic and discriminator networks follow that same architecture, although they do not share parameters.
The inputs to both networks correspond to 256x140 RGB images from the three frontal cameras.
When stacked, these images yield an input with 9 channels, 
%140 pixels height and 256 pixels width 
that is fed to the convolution block (Fig.~\ref{fig:nn}).
The other continuous input is the car's linear velocity, which is concatenated with the 8-dimensional input from the trajectory as well as to the flattened feature vector from the last convolutional layer. The discriminator has an additional continuous input for the action.
%and have an input of three RGB images from the forward facing cameras installed on the vehicle on the convolution block. And both receive as input directly on the fully connected block the vehicle speed, the next target on the vehicle coordinate system and the high-level command.

%On the discriminator network the linear inputs are concatenated with the policy action. The network output goes through different activation fuctions for the discriminator training and the actor critic reward function. For the former the activation function is tanh and for the last is a sigmoid function. That is a implementation of reward shapping following the architecture on \cite{Zhang_2020}.

For the actor-critic network, three outputs compose the last layer of the network: a linear unit for the value $V(s)$, 
a $\tanh$ unit for the steering wheel action, and a sigmoid unit for the throttle action, restricting the outputs to the valid domain of these commands \cite{li2017infogail}.

%--- Achei muito fraquinho a visualização para mostrar
% \begin{figure*}[htp]
%     \centering
%     \includegraphics[width=5.5cm]{vector_fields/early_eval.eps}
%     \includegraphics[width=5.5cm]{vector_fields/late_eval.eps}
%     \label{fig:trainVectorFieldImages}
% \end{figure*}

% The final inputs to the agent network are the three images from the frontal cameras stacked to generate an image with 9 channels, which originate from the three RGB channels for each image. 
% And a vector of nine continuous inputs ahead in the dense part of the network consisting of concatenated one input of linear velocity module, two inputs of the next trajectory point in the car reference and the next trajectory high level command as a one hot encoded vector with the six options of command.
% \subsubsection{Networks inputs}
% Comment: mention that the last input is not so necessary

% Though the agent receives input to follow a generic trajectory the experiment is conducted for a fixed trajectory. 

%EA: comentei acima, por ja estar na conclusao isso

%\subsection{Notes on Learning}
% The agent learning process is based on the use of a 
% non deterministic policy, to calculate actions probabilities. To make the policy non deterministic was employed a Gaussian distribution with mean on the policy network output and fixed standard deviation. The same approach as \cite{li2017infogail} that also built an experiment around a 3d vehicle simulator.
\subsubsection{Non deterministic policy}
The agent learning process is based on the use of a 
stochastic policy to calculate action probabilities. 
This is achieved by using the Gaussian distribution, whose mean is predicted by the policy network, and the standard deviation is fixed to a predefined value \cite{li2017infogail}. This was necessary because a variable entropy was shown to be not suitable:
the agent with a high entropy is easily disturbed on sensitive moments like a turn, whereas there is not enough exploration during turns if the entropy is too low.
%The same approach as \cite{li2017infogail} that also built an experiment around a 3d vehicle simulator.
% \eric{integrar o texto acima....}

%The use of a fixed standard deviation, diverges from the variable entropy approach on the original Proximal Policy Optimization \cite{ppoOriginal} algorithm proposal, that starts with a higher entropy to improve the agent exploration, and uses a entropy coefficient on the policy loss to guarantee a grade of exploration.

%The CARLA environment is very sensitive to the entropy, if the entropy is high the agent fails the trajectory cause of disturbs on control on sensitive moments like a turn. And if the entropy is too low when the agent do not explore enough to learn how to execute the turns.

%-EA: comentei acima, por não conseguir tirar nenhuma conclusão aparente

\section{Experiments}
\label{experiments}
%traj longa
%760 pontos densa
%20 pontos esparsa
%traj curta
%80 pontos densa
%4 pontos esparsa
\begin{figure*}[htp]
    \centering
    \includegraphics[width=4.0cm]{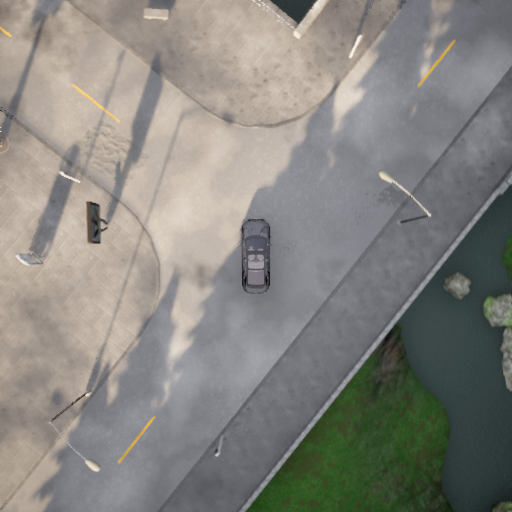}
    \includegraphics[width=4.0cm]{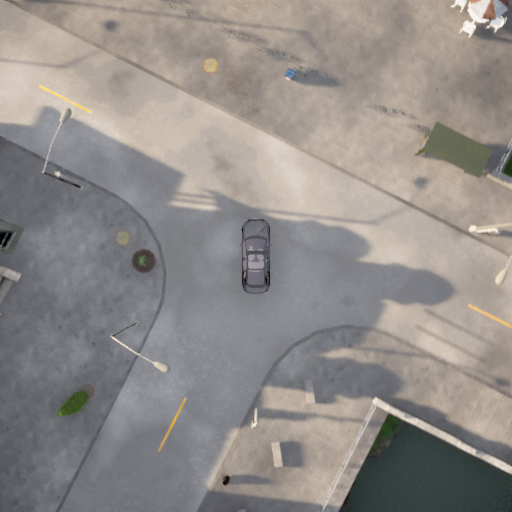}
    \includegraphics[width=4.0cm]{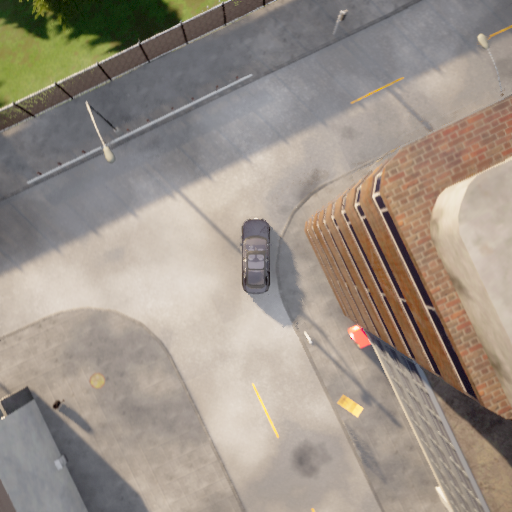}
    \includegraphics[width=4.0cm]{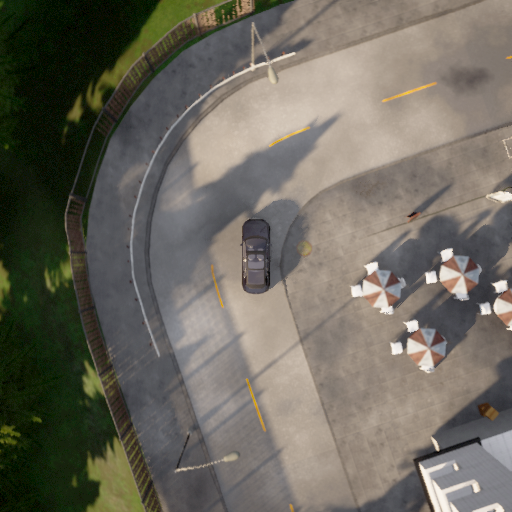}
    \caption{
    \bb{The top-down view of the simulation with the car in the center and making a turn for the long route. Each picture shows one of the four possible turns, from left to right: left, left, right, and right turns.
    }
    }
    \label{fig:trajImages}
\end{figure*}

\bb{The  learning  navigation  experiments  are  inspired  on  the CARLA  Leaderboard  evaluation  platform and consists of navigating autonomously on two setups: a short route of 100 meters and one turn (setup 1); and a long route of $2,500$
meters and four turns (setup 2). The long route
was chosen from the ones available in the CARLA Leaderboard \cite{leaderboard_2020}. The short one corresponds to the first 100 meters of the long route.
}

A top down image from the simulator presenting each turn from the trajectories is displayed on (Fig.~\ref{fig:trajImages}).

% \eric{Organize the experiments in two sections: Short trajectory and long trajectory}
% \eric{Save another section (that comes first) that is common to both experiments}
% \eric{always refer to the Figures on the text}

% \subsection{Demonstration}
\subsection{Dataset}

The expert dataset is built using a deterministic agent that navigates using a dense point trajectory
and a classic PID controller \cite{chen2019learning}. 
\bb{While a dense point trajectory provide many points at a finer resolution, a sparse point trajectory is made of considerably less points to follow, providing just a sense of the right direction to the agent. Thus, the former is used to generate training data by the expert, while the latter is used by the agent for more high-level directions.
For instance, the first setup (the short route) considers 80 and 4 points for the dense and sparse trajectories, respectively.
The second setup (the long route) uses
760 points in the dense trajectory, and 20 points in the sparse one.
}%That agent is the same one used to build a dataset for the Learning by Cheating \cite{chen2019learning} paper where in the first stage the agent learns using behavior cloning.

\bb{For both setups, 10 complete trajectories were recorded at 10 hertz, i.e., 10 observation-action pairs per second were generated.
For the short route, those trajectories correspond to 5 minutes of driving as if in a real scenario,
totalling $3,000$ training samples (4GB of uncompressed data).
For the long route, those trajectories correspond to half an hour of driving, totalling $18,000$ training samples (30GB of uncompressed data). 
}
\subsection{Training}
\bb{The training was performed using ten parallel actors in a synchronous way, each one running its own CARLA simulator. An eleventh CARLA simulator was also run for evaluation purposes.
}

\bb{In a simulation, every episode starts with the vehicle at zero speed on a particular initial point.
The episode ends at every infraction, collision or lane invasion and a new episode starts with the vehicle initially located where the infraction occurred with 90\% chance.
%for that the trajectory is always monitored from the environment using a route planner.
With 10\% chance, the location in the trajectory is randomly chosen, in order to diversify the experience for each policy update.
}%  at every infraction there is also a chance of one to ten of episode restart from a trajectory random point instead of the last point from the trajectory that the agent crossed.

\bb{
For the short (long) route, 240 (720) environment interactions or timesteps are recorded for every actor and then the resulting training set of $2,400$ ($7,200$) samples is used to train the parametrized policy in a central computer using (\ref{eq:bcgail}) as loss function. Thus, the episode does not have to end for a policy update to happen.
Notice that any one of the ten actors can be interacting with the environment in different parts of the trajectory at a certain moment.
Other hyperparameters can be seen in Table~\ref{tab:hyperparams}. For instance, the standard deviation of the Gaussian distribution ($\sigma_1$ for steer and $\sigma_2$ for throttle) for the policy is fixed to a predefined value. Thus, the output of the policy network only affects the mean of the distribution.
}

\bb{A behavior cloning (BC) agent is also trained for comparison, using the same dataset available for the GAIL agents,
but 70\% of the samples are used for training, while 30\% for  validation. 
The dataset is not augmented using random rotations or shifting, so that the techniques are compared on their sample efficiency on the same expert trajectories.
The BC agent is trained using the loss function from (\ref{eq:bc}), and an ADAM optimizer with a learning rate of $3.0 \times 10^{-4}$. The network with best validation error is then evaluated in the simulation experiments.
}
%The agent is trained for a hundred epochs, and the epoch with best validation score is recorded to generate the agent used for evaluation on the environment. 

% 7200 env. interactions / 10
% ADAM, 128 minibatch discriminator
% PPO, 8 minibatch  
%

\begin{table}[]
    \centering
        \caption{Hyperparameters for training }    
        \label{tab:hyperparams}
    \begin{tabular}{lcc}
    \hline
        &  Short Route              &  Long Route  \\ \hline
    Parallel environments ($N$)         & $10 $                       & $10$                       \\    
    Adam step size (lr)                 & $1.0 \times 10^{-4}$    & $1.0 \times 10^{-4}$   \\
    Number of PPO epochs (K)            & $4 $                      & $4$                      \\
    Mini-batch size (m)                 & $300  $                     & $900$                      \\
    Discount ($\gamma$)                 & $0.99$                     & $0.99$                    \\
    GAE parameter ($\lambda$)           & $0.95   $                  & $0.95 $                   \\
    Clipping parameter ($\epsilon$)     & $0.1  $                    & $0.1  $                   \\
    Value Function coefficient ($c_{1}$)& $0.5 $                     & $0.5  $                   \\
    Entropy coefficient ($c_{2}$)       & $0.0 $                     & $0.0  $                  \\
    Timesteps per epoch (T)               & $2400 $                    & $7200  $              \\
    Log Standard Deviation Steer ($\sigma_1$) & $ -2.0 $                 & $ -2.0  $
    \\
    Log Standard Deviation Throttle ($\sigma_2$)     & $ -3.2 $                 & $ -3.2 $
    \\
    \hline
    \end{tabular}
    
\end{table}

\section{Results}
\label{results}
%-EA: O texto comentado abaixo pode ser aproveitado em outra seção

% The training was performed using ten parallel Carla simulators and an eleventh for evaluation purpose. For that was used a three nodes GPU cluster, with two GTX 1050 and one GTX 1060 distributed on two nodes dedicated for the simulators and one main node dedicated to train the network with one RTX 3060.

% On that setup each training corresponded to 8 hours of computing on the three nodes for the short experiment and 30 hours for the long experiment.
% % each step on the simulator corresponds to 0.1 second of driving

% And 19 worth of driving on the simulator for the complete experiment, and 3 hours to learn with the gail, one hour and half for the bc regularized gail.

% For the long were 50 hours worth of driving for the complete experiment and 30 hours for the gail to learn and 15 hours to the bc regularized gail.

% The experiment was made using two trajectories, a short trajectory with only one intersection turn and a long trajectory with two intersection turns, and two lane curves. The short trajectory is a subset of the long one. And the long trajectory is sampled from the trajectories defined on Carla Leaderboard.

In order to evaluate the performance of agents, the reward or score metrics is defined as the number of crossed points from the dense trajectory, representing how much of the trajectory the agent has completed without any mistake. Thus, a maximum reward is equivalent to total number of points in the dense trajectory of the particular route.

%The reward metrics is generated by the environment for evaluation purpose. To generate the metric the environment monitors the route completion using an instance of the route planner created from the dense point trajectory, so that the reward metrics reflects the total of points from the dense trajectory that the agent crossed during the episode.
% \eric{Explain this reward metric better}

\begin{figure}[]
\centering
\includegraphics[width=.4\textwidth]{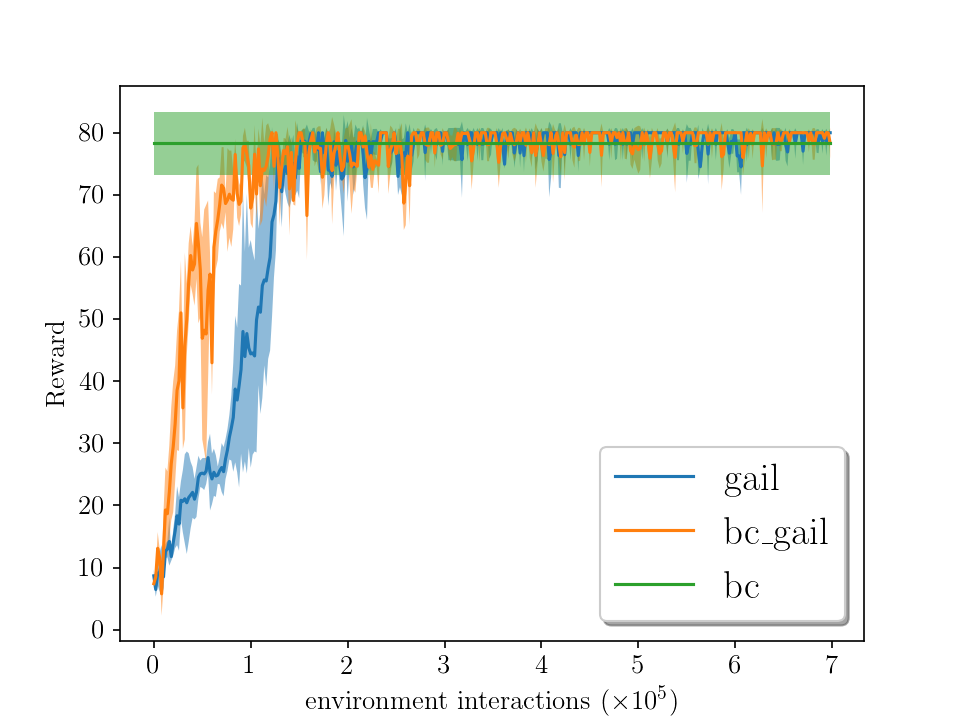}
\includegraphics[width=.4\textwidth]{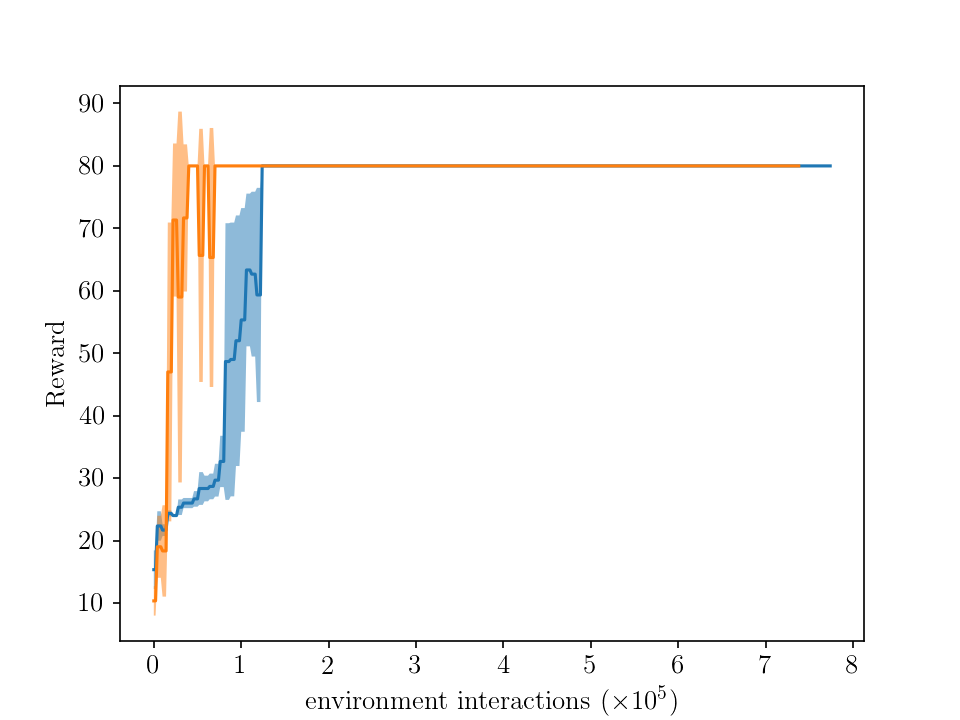}
\caption{Average rewards vs environment interactions during training in the short route (setup 1). 
For each method (GAIL and GAIL with BC), the average performance of three runs (i.e, three agents trained from scratch)
is shown with a stochastic policy (top plot) and a deterministic policy (bottom plot). The shaded area represents the standard deviation.
The behavior cloning (BC) agent attains an average reward of 78.3 for ten episodes, while the maximum is at 80, achieved by both GAIL and GAIL augmented with BC.
\eric{Descrever figura}
\eric{Essa figura plota uma média móvel? Os outros artigos geralmente plotam a média, uma suavização, correto?}
}
\label{fig:short_reward}
\end{figure}

\begin{figure*}[]
\centering
\includegraphics[width=.44\textwidth]{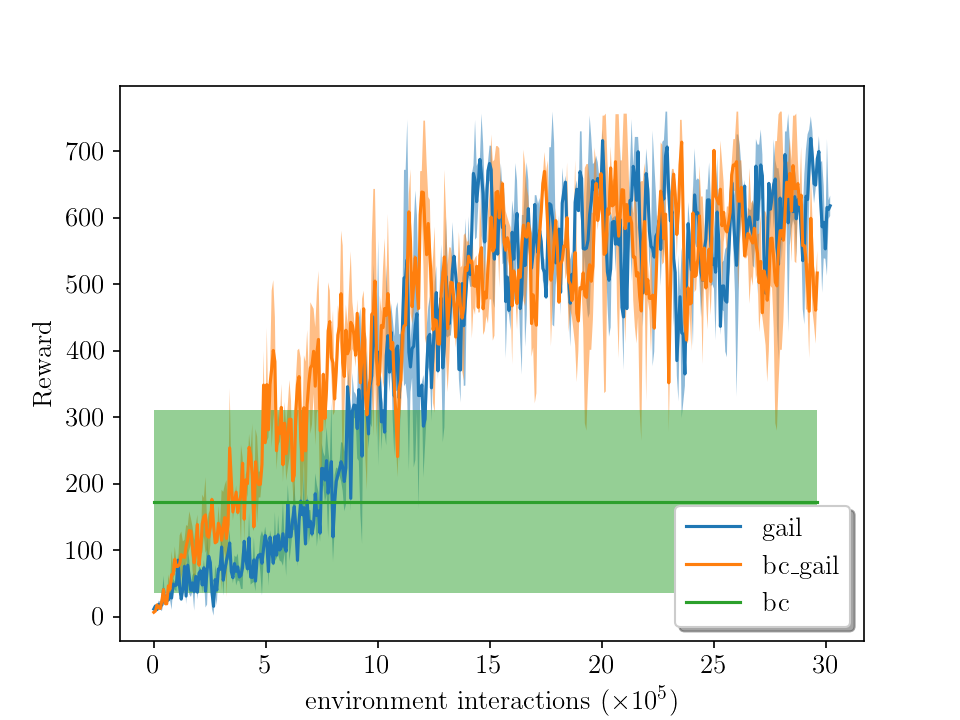}
\includegraphics[width=.44\textwidth]{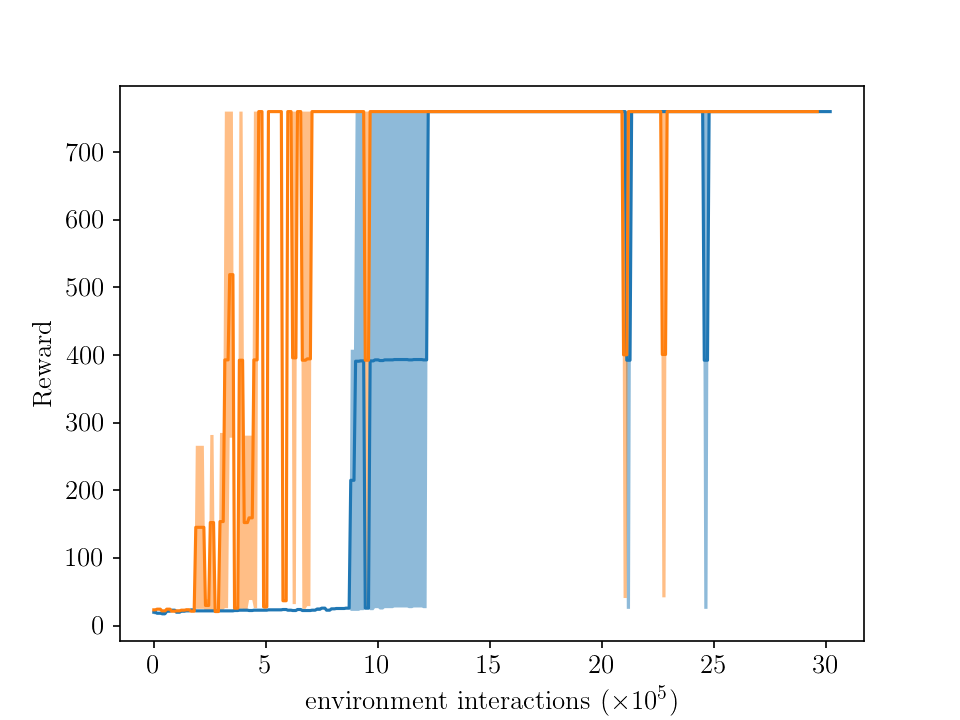}
\caption{
Average rewards vs environment interactions during training in the long route (setup 2). For each method (GAIL and GAIL with BC), the average performance of two runs (i.e, two agents trained from scratch) is shown with a stochastic policy (left plot) and a deterministic policy (right plot).
The shaded area represents the standard deviation.
The behavior cloning (BC) attains 
an average reward of 173.6 for ten episodes, while the maximum is at 760, achieved by both GAIL and GAIL augmented with BC.
% \eric{Descrever figura (média dos últimos 10 o que? média em termos de environment interactions?)
% }
% \eric{Essa figura plota uma média móvel? Os outros artigos geralmente plotam a média, uma suavização, correto?}
}
\label{fig:long_reward}
\end{figure*}

The learning performance of both GAIL and GAIL augmented with BC (BC\_GAIL) can be seen on Figures \ref{fig:short_reward} and \ref{fig:long_reward} for the short and long routes, respectively. In the first setup, BC\_GAIL is able to converge significantly faster than GAIL, achieving maximum reward of 80, slightly higher than just behavior cloning.
On the second more challenging setup which consists of four turns, the learning takes considerably more time. 
On average, the agent by BC\_GAIL was able to complete the route without any mistake much earlier than the GAIL agent, also showing early fast improvement of the policy. This is possible due to the strong influence of the BC term in the loss function in the early part of the training process.
Notice that an agent trained only by BC is not able to solve this task (achieved only a reward of 173.8) by training only on the same dataset as GAIL was trained.
In addition, %in simulations taking approximately $40 \times 10^5 $ environment interactions, which is more than double the time in Fig. \ref{fig:long_reward}, 
after $15 \times 10^5$ environment interactions, we can observe that the average reward stabilizes between 500 and 700 for the stochastic policy of both GAIL and BC\_GAIL. %(now shown in the plot).
The spikes seen in Fig. \ref{fig:long_reward} can be caused by random actions of a stochastic policy which can lead to forgetting of some already acquired skills (such as turning at an intersection) or skills that are not well formed yet. For instance, the agent can learn to make a turn at some point and, after some iterations, fail to repeat that behavior, causing a sudden drop of the reward. This happens because turning is a difficult skill to learn, while the reward is proportional to the traveled distance.

% - curvas
% - esquecimento
% - aleatoriedade

%8071

\begin{figure}[]
    \centering
    \includegraphics[width=8.8cm]{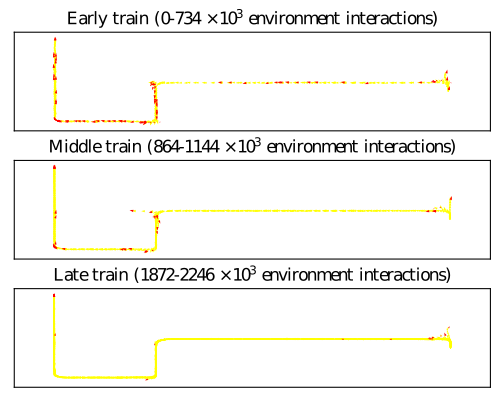}
    \caption{
    \bb{The vehicle's trajectory, in yellow, for the long route during different moments of the training process. In the early training iterations, errors, marked in red color, are common. As training proceeds, less and less mistakes happen.
    The trajectory starts at the right side, heading North, and ends at the left side, also heading North.
    }
    }
    \label{fig:trainVectorFieldImages}
\end{figure}

The trajectory of the agent for the long route can be viewed
in Fig.~\ref{fig:trainVectorFieldImages}. It shows the early mistakes in red color made by an BC\_GAIL agent in the topmost plot. As training proceeds, less and less mistakes are made as it can be noticed in the remaining plots. 

\section{Conclusion}
\label{conclusion}
In this work, we have proposed a GAIL-based architecture for end-to-end autonomous driving in urban environments. 
Despite the known difficulties and learning instabilities of generative adversarial networks,
both GAIL and GAIL augmented with BC were able to converge and generate agents able to complete the whole trajectory without mistakes, with the latter able to quickly find a suitable policy when compared to the former.
Both of them surpassed Behavior Cloning in performance, which was not able to generate an agent even capable of making more than one turn on average in the long route.

Although the trajectories were fixed beforehand, the architecture is general enough to allow for variable routes, i.e., an agent that can change course in real time depending on a dynamic route,
which will be tackled as future work.
We also plan to investigate risky scenarios with rare events, and which loss functions or models could be of use to handle those important settings. This is where reinforcement learning approaches can make a difference as BC depends on static collected data which might not be representative of all possible real-world scenarios.

\bibliographystyle{IEEEtran}
\bibliography{IEEEabrv,conference}

\end{document}